\documentclass[runningheads]{llncs}
\usepackage[T1]{fontenc}
\usepackage{graphicx}

\usepackage{xcolor,colortbl}

\usepackage{orcidlink}
\usepackage{academicons}

\usepackage{tikz}
\usepackage{amsmath}
\def\checkmark{\tikz\fill[scale=0.4](0,.35) -- (.25,0) -- (1,.7) -- (.25,.15) -- cycle;} 
\usepackage[utf8]{inputenc}
\usepackage{float}
\usepackage{indentfirst}
\usepackage{multirow}
\usepackage{longtable}
\usepackage{subcaption}
\usepackage{tabularray}
\usepackage{pdflscape}
\usepackage{hyperref}

\usepackage[normalem]{ulem}

\makeatletter
\newcommand{\printfnsymbol}[1]{%
  \textsuperscript{\@fnsymbol{#1}}%
}
\makeatother

\newcommand{\myparagraph}[1]{\vspace{0.5em}\noindent\emph{#1}}
\newcommand{\toch}[1]{{#1}}

\newcommand{\todo}[1]{}

\newcommand{\todel}[1]{}

\begin{document}
%

\title{The Role of Explanation Styles and \\Perceived Accuracy on Decision Making in \\Predictive Process Monitoring} 

%

%
%

%

\author{
Soobin Chae\thanks{contributed equally as co–first authors}
\and
Suhwan Lee\printfnsymbol{1}~\orcidlink{0000-0001-8089-0960} 
\and
Hanna Hauptmann~\orcidlink{0000-0002-6840-5341}
\and\\
Hajo A. Reijers~\orcidlink{0000-0001-9634-5852} 
\and
Xixi Lu~\orcidlink{0000-0002-9844-3330}
}
\authorrunning{S. Chae et al.}
\titlerunning{Explanation Styles and Perceived Accuracy in PPM Decision-Making}
%
\institute{Utrecht University, Utrecht, The Netherlands\\
\email{\{s.lee, h.j.hauptmann, h.a.reijer, x.lu\}@uu.nl}}
\maketitle              
\begin{abstract}
Predictive Process Monitoring (PPM) often uses deep learning models to predict the future behavior of ongoing processes, such as predicting process outcomes. 
While these models achieve high accuracy, their lack of interpretability undermines user trust and adoption.
%
Explainable AI (XAI) aims 
to address this challenge by providing the reasoning behind the predictions. However, current evaluations of XAI in PPM focus primarily on \emph{functional metrics} (such as fidelity), overlooking \emph{user-centered} aspects such as their effect on \emph{task performance} and decision-making.
This study investigates the effects of explanation styles (feature importance, rule-based, and counterfactual) and perceived AI accuracy (low or high) on decision-making in PPM. We conducted a decision-making experiment, where users were presented with the AI predictions, perceived accuracy levels, and explanations of different styles. Users’ decisions were measured both before and after receiving explanations, allowing the assessment of objective metrics (Task Performance and Agreement) and subjective metrics (Decision Confidence). Our findings show that \toch{perceived accuracy and explanation style have a significant effect.} 


\end{abstract}
%
%

%

\section{Introduction}\label{sec:intro}

Predictive Process Monitoring (PPM) is a set of techniques that uses historical event logs to predict the future behavior of ongoing processes using machine learning~\cite{DBLP:books/sp/22/FrancescomarinoG22}. These techniques enable predictions across various dimensions, such as forecasting the remaining time of a case, the next process step, and outcomes of ongoing processes~\cite{DBLP:conf/bpm/RizziFM20}. 
By integrating these predictive capabilities into decision support systems, PPM can empower organizations to make data-driven decisions, enhancing efficiency and productivity across various domains, including healthcare, finance, and business operations.

Recent PPM techniques often use complex predictive models, such as deep learning, to achieve high prediction accuracy. 
This improved accuracy often comes at the expense of interpretability, as these ``black box'' models are inherently difficult for users to understand. 
%
This lack of interpretability and transparency can lead to hesitancy among users—such as decision-makers and process owners—to trust and adopt such systems~\cite{DBLP:journals/electronicmarkets/HammKCW23}. 
This issue is particularly critical in the PPM domain, where predictions can directly influence crucial decisions, impacting individual cases and potentially entire workflows~\cite{DBLP:journals/algorithms/LeeCK22}. 

Explainable AI (XAI) has emerged as a promising solution to address this interpretability challenge. XAI aims to explain the ``how'' and ``why'' behind predictions while preserving the strong predictive performance of complex models~\cite{DBLP:journals/inffus/ArrietaRSBTBGGM20}. In the PPM domain, recent advances have adapted XAI techniques, such as LIME~\cite{DBLP:conf/kdd/Ribeiro0G16} and DiCE~\cite{DBLP:conf/fat/MothilalST20}, to explain process predictions~\cite{DBLP:journals/eaai/GalantiLMNMSM23,DBLP:conf/icpm/HsiehMO21,DBLP:conf/ecis/MehdiyevF20,DBLP:journals/tist/VerenichDRMT19} or evaluate predictive models~\cite{DBLP:conf/bpm/RizziFM20}. By offering interpretable insights, these techniques have the potential to increase user trust and foster the adoption of PPM technologies.

Despite these advancements, user evaluations of XAI explanations remain scarce, particularly very few studies focus on the \emph{effectiveness} of explanations rather than their \emph{understandability}~\cite{DBLP:journals/corr/Doshi-VelezK17,DBLP:conf/atal/AnjomshoaeNCF19}. 
Much of the current XAI evaluation focuses on \emph{functional} metrics (e.g., fidelity) rather than the \emph{user-centered} effects of explanations. Many questions remain unanswered. For example, how do these explanations affect users' \emph{task performance}, \emph{agreement} with AI predictions, or \emph{confidence} in their decisions? 

In the PPM domain, only a few studies~\cite{DBLP:journals/eaai/GalantiLMNMSM23,DBLP:journals/corr/abs-2202-07760} have conducted user-centered evaluations, focusing on a single XAI technique and focusing on usability or understandability. 
There remains a critical gap in understanding the effects of different explanation styles and perceived AI accuracy on decision-making, such as task performance, agreement, and decision confidence.





In this paper, we aim to address this gap through an empirical, user-centric evaluation focusing the effectiveness. We investigate these two factors, namely (1) \emph{explanation styles} and (2) \emph{perceived AI accuracy}.  
Our motivation for focusing on these two factors is the following. 
Explanation styles vary in their logical structure and how they present information to users, influencing users' reasoning processes~\cite{DBLP:journals/ai/WaaNCN21,DBLP:journals/tiis/CauHST23}. 
\toch{Perceived AI accuracy, on the other hand, impacts user trust and reliance, with prior research suggesting a relation between perceived correctness and the information users incorporate in decision-making~\cite{DBLP:journals/ai/KennyFQK21,cecil2024explainability}.}
%
%
We investigate the following research question: \emph{How do (1) different explanation styles and (2) the perceived accuracy of AI predictions affect decision-making in PPM?} 

%
%

Our evaluation was conducted in a decision-making context where participants determine whether to accept or reject loan applications based on AI predictions.
Specifically, we trained a ``black-box'' predictive model on the loan application log \toch{from BPIC 2017~\cite{bpic17}.} and used XAI techniques representing three distinct styles to generate corresponding explanations. To create the survey, we randomly selected \emph{four cases} (two predicted correctly and two incorrectly) along with their predictions and explanations. 
The study involved 179 participants, divided into independent groups. Each participant received an introduction, detailed descriptions of the task, and then the four selected cases (shown sequentially). For each case, participants were asked to make decisions both before and after being presented with the explanations. To evaluate the impact of these explanations, we measured a combination of objective metrics (\emph{Task Performance} and \emph{Agreement}) and a subjective metric (\emph{Decision Confidence}). 
Our findings contribute to a deeper understanding of how users interact with XAI explanations, providing insights for designing more effective decision-support systems in PPM.

The remainder of this paper is structured as follows: Section~\ref{sec:relatedwork} discusses the current XAI techniques in the PPM domain and their evaluation methods. 
Section~\ref{sec:research-method} presents the research method. 
Section~\ref{sec:results} covers the evaluation of the developed system. 
Finally, Section~\ref{sec:conclusion} summarizes our key findings and discusses future work.

\section{Related Work}\label{sec:relatedwork}

Recently, explainability approaches have also been adapted and investigated in the field of PPM~\cite{DBLP:conf/icpm/HsiehMO21,DBLP:conf/caise/HundoganLDR23,DBLP:conf/bpm/BuligaVGLDFGR24,DBLP:conf/caise/BuligaFGM23}
Only a few studies in the PPM domain are concerned with evaluating the explanations generated with XAI techniques. As the importance of explaining PPM results gains recognition, the evaluation of these explanations is expected to gain more interest. 
Explainability evaluation can be categorized into three types~\cite{DBLP:journals/corr/Doshi-VelezK17}: \toch{Function-grounded, Application-, and Human-evaluations}. In Table \ref{evalexp}, relevant studies are classified based on their evaluation approaches and whether they compare XAI methods.
 
Function-grounded evaluation assesses explanations based on their inherent characteristics, such as stability and fidelity. \emph{Stability} refers to the consistency of explanations generated for the same data sample under identical conditions~\cite{visani2022statistical}. They proposed metrics to evaluate the stability of the top-K feature subset and their respective weights across multiple explanations for certain process instances~\cite{DBLP:conf/icsoc/Velmurugan0MS21}. \emph{Fidelity}, on the other hand, pertains to the ability of XAI methods to accurately mimic the behavior of the explained ML model in the vicinity of the explained process instance. 
Velmurugan et al.~\cite{DBLP:conf/caise/Velmurugan0MS21} introduced an approach to evaluate local XAI methods for their internal fidelity, which compares the decision-making process of the explainer proxy model with the explained black-box model. 
%
Among the Function-grounded evaluation studies in the PPM domain, Stevens et al.~\cite{stevens2023explainability} and Elkhawaga et al.~\cite{elkhawaga2024should} proposed evaluation approaches for various XAI methods applied to PPM results. 
Stevens et al. introduced four out-of-the-box metrics from relevant XAI evaluation research and applied them to different attributes of process mining data. 
Elkhawaga et al.~\cite{elkhawaga2024should} proposed an approach for evaluating the consistency of XAI methods. While useful, we lack insights into how these explanations affect user's decision-making. 

Only two studies~\cite{DBLP:journals/corr/abs-2202-07760,DBLP:journals/eaai/GalantiLMNMSM23} focused on \emph{user evaluations}, specifically \emph{application-grounded} evaluation. Rizzi et al.~\cite{DBLP:journals/corr/abs-2202-07760} are among the first to investigate whether users understand the explanation plots. Instead of applying actual XAI methods, they generate three levels of different plots in event, trace, and event log levels. While the study involved participants from both the PPM and ML fields, comprehension and usage levels of explanations varied based on domain knowledge and experience. However, the study relied on qualitative evaluation with a limited number of participants without employing a consolidated user-interface evaluation methodology. Galanti et al.~\cite{DBLP:journals/eaai/GalantiLMNMSM23} proposed a framework to evaluate the understanding and comfort level of process analysts with results from an explainable predictive monitoring framework. Their evaluation focused on accuracy in task execution, perceived task difficulty, usability, and user experience dimensions. Regarding the comparison of XAI methods, no user study compares the effectiveness of different XAI styles in user evaluation in the PPM domain.

\begin{table}[tb]
\centering
\caption{Explainability Evaluation Approaches in PPM}
\label{evalexp}
\resizebox{\columnwidth}{!}{%
\begin{tabular}{lllcccc}
\hline
\textbf{Year} & \multicolumn{1}{c}{\textbf{Ref.}} & \multicolumn{1}{c}{\textbf{Evaluation Metric}} & \textbf{\begin{tabular}[c]{@{}c@{}}Function \\ grounded\end{tabular}} & \textbf{\begin{tabular}[c]{@{}c@{}}Application \\ grounded\end{tabular}} & \textbf{\begin{tabular}[c]{@{}c@{}}Human\\ Evaluation\end{tabular}} & \multicolumn{1}{l}{\textbf{\begin{tabular}[c]{@{}l@{}}XAI Methods \\ Comparison\end{tabular}}} \\ \hline
2020 & Velmurugan et al.~\cite{DBLP:conf/caise/Velmurugan0MS21} & Fidelity & \checkmark &  &  & \checkmark \\ \hline
2021 & Velmurugan et al.~\cite{DBLP:conf/icsoc/Velmurugan0MS21} & Stability & \checkmark &  &  & \checkmark \\ \hline
2022 & Huang et al.~\cite{DBLP:journals/corr/abs-2202-12018} & \begin{tabular}[c]{@{}l@{}}Fidelity\\ Domain Knowledge\end{tabular} & \checkmark &  &  &  \\ \hline
2023 & Stevens et al.~\cite{stevens2023explainability} & \begin{tabular}[c]{@{}l@{}}Parsimony\\ Functional Complexity\\ Importance Ranking Correlation\\ Level of Disagreement\end{tabular} & \checkmark &  &  & \checkmark \\ \hline
2024 & El-Khawaga et al.~\cite{elkhawaga2024should} & Consistence & \checkmark &  &  & \checkmark \\ \hline
2022 & Rizzi et al.~\cite{DBLP:journals/corr/abs-2202-07760} & Understandability &  & \checkmark &  &  \\ \hline
2023 & Galanti et al.~\cite{DBLP:journals/eaai/GalantiLMNMSM23} & \begin{tabular}[c]{@{}l@{}}Accuracy\\ Perceived Task Difficulty\\ Usability\end{tabular} &  & \checkmark &  &  \\ \hline
\end{tabular}
}
\end{table}

We identified two key gaps that motivate our work.  
First, there are limited comparative user evaluations in the PPM domain. Existing user evaluations of XAI techniques for PPM primarily focus on individual XAI methods or qualitative evaluations with limited generalizability. Second, there is a lack of research that studies the effect of different explanation styles on decision-making, not to mention the perceived AI accuracy. Existing research primarily focuses on the understandability and usability of a single explanation style. There is no systematic investigation into how different explanation styles and the accuracy of a prediction model affect a user's decision process.

Outside the area of PPM, there have been studies that show how different explanation styles and AI accuracies can impact the user's decision confidence, task performance, and agreement. In~\cite{DBLP:conf/iui/CauHST23}, two explanation styles are studied in the domain of stock market trading. The study shows that both rule-based and feature-importance explanations improved task performance, but only in cases of high AI accuracy. A follow-up paper~\cite{DBLP:journals/tiis/CauHST23} on text and image classification tasks showed that the domain of the prediction model can change the way explanation styles impact the user's task performance, indicating that a new assessment in the PPM domain is required. 
Both papers further propose expanding such comparisons by including counterfactual examples as an explanation technique.

Given these gaps, this paper aims to conduct an empirical user evaluation to investigate the effect of different explanation styles and perceived AI accuracy on decision-making.  


\section{Research Method}\label{sec:research-method}
In this section, we outline the research methods for conducting the empirical user evaluation.
The section is organized as follows: we introduce the evaluation objectives, describe the experimental design, define the hypotheses and analytical approach, and detail the overall experimental procedure.

\subsection{Evaluation Objectives}\label{evob}
The objective of this study is to examine the \emph{effect} of explanation styles and the \emph{perceived AI accuracy} on decision-making within the context PPM. Specifically, the evaluation is structured around the following objectives:


\begin{enumerate}
    \item To investigate the effect of perceived AI accuracy on task performance, agreement with AI predictions, and decision confidence. 
    \item To investigate the effect of explanation styles on task performance, agreement, and decision confidence by comparing results before and after the provision of explanations.
\end{enumerate}

\todel{The first objective examines how perceived AI accuracy, categorized as high or low, influences decision-making in terms of task performance, agreement, and decision confidence. Understanding these influences will help categorize decision-making situations based on prediction accuracy. The second objective focuses on evaluating the effects of various explanation styles— Feature importance-based, Rule-based, and Counterfactual-based—on the aforementioned effectiveness metrics. The aim is to determine which styles best support informed decision-making, effectively persuade users to take specific actions and enhance their confidence in their decisions. The third objective involves assessing how decision-making metrics change after explanations are provided. The aim is to identify which explanation style influences the most in affecting task performance, agreement, and decision confidence. The final objective is to explore the reasons for differences in explanation effectiveness, focusing on external subjective factors such as users' backgrounds and satisfaction levels. This will help us to gain insights into the factors influencing the impact of explanations.}

\subsection{Experimental Design}
\begin{figure}[tb]
    \centering
    \includegraphics[width=\textwidth]{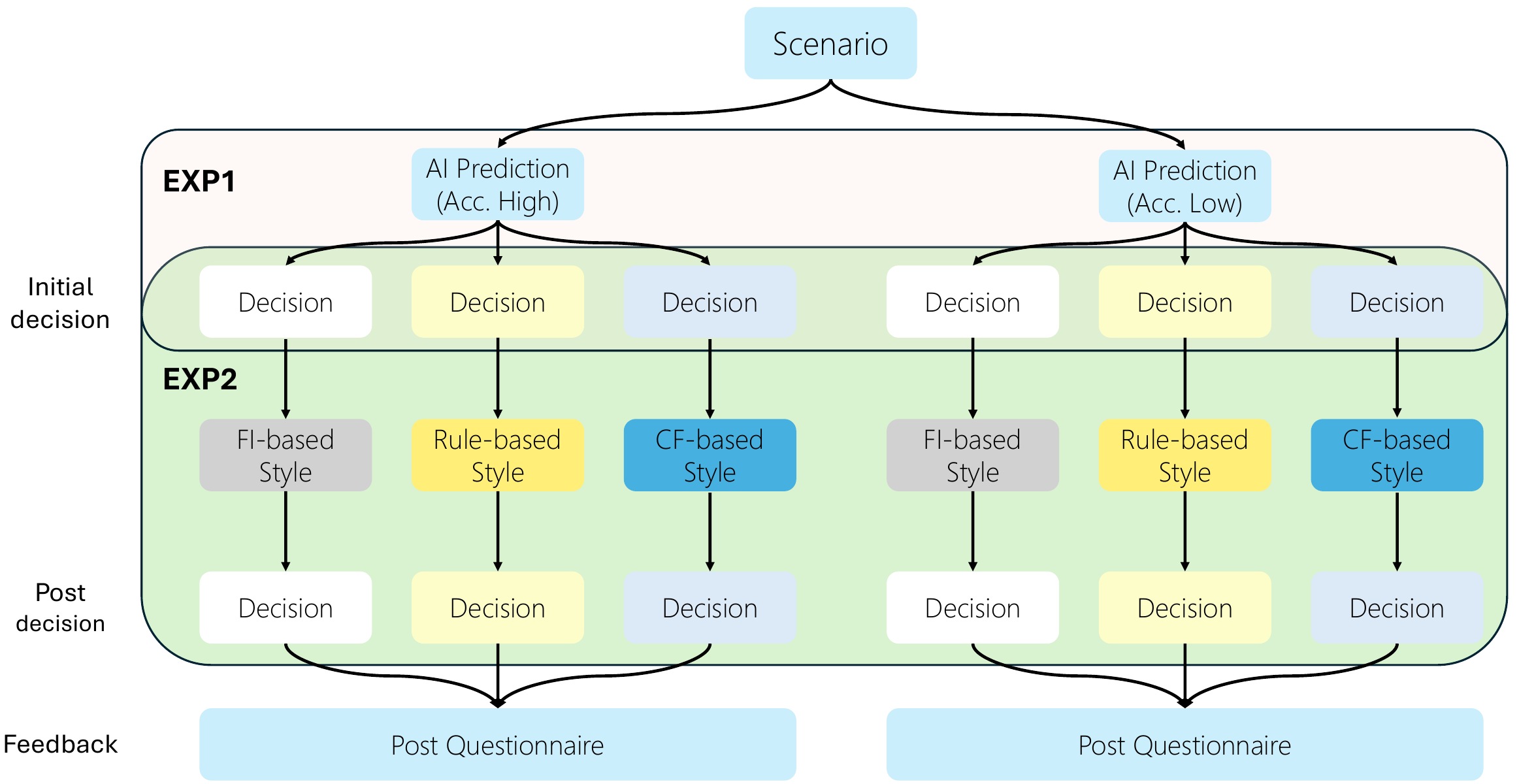}
    \caption{Experiment Setting}
    \label{fig:experimentsetting}
\end{figure}

The experiment was designed to address the evaluation objectives outlined in the previous section. As illustrated in Figure~\ref{fig:experimentsetting}, a two-phase decision-making setup was implemented to examine the effects of perceived accuracy and explanation styles. 
%
\toch{First, participants were randomly assigned into two groups;
the perceived AI accuracy was manipulated depending on the group, i.e., the high-accuracy group was shown a 96\% accuracy, whereas the low-accuracy group was shown a 63\% accuracy.} This division allowed us to assess the effect of perceived accuracy on participants' initial decisions (EXP1). 
Within each group, participants were further assigned to one of three commonly used explanation-style subgroups: Feature Importance (FI), Rule-based, or Counterfactual (CF)~\cite{raufi2024comparative}, thus in total, 6 independent groups. 
This setup facilitated comparisons of pre- and post-explanation decision-making across explanation styles.

Each participant is randomly assigned to follow one of these six branches in Fig.~\ref{fig:experimentsetting} (e.g., High-accuracy and FI-based style) and sees four cases sequentially: two cases with correct predictions and two cases with incorrect predictions. For each case, participants' \emph{decisions}, i.e., \emph{whether they agree with or are against the AI decision}, were recorded at two points: \textbf{before} explanations were provided (initial decisions) and \textbf{after} receiving the assigned explanation style (post-explanation decisions). This design enabled both within-group (pre- versus post-explanation) and cross-group (accuracy and explanation style) comparisons.

Two \emph{independent variables} were recorded in this experiment:
    (1) \textbf{Perceived AI Accuracy}: high (96\%) or low (63\%), determining the accuracy level participants were exposed to;
    (2) \textbf{Explanation styles}: the assigned explanation style, categorized as Feature Importance, Rule-based, or Counterfactual.
%

To evaluate the effect of explanation styles and perceived AI accuracy on decision-making, the following three \emph{dependent variables} were measured for each participant across the four cases, both \emph{before} and \emph{after} explanations:
   \begin{enumerate}
    \item \textbf{Task performance}: The number of correct decisions participants make across the four cases, measured separately before and after.
    \item \textbf{Agreement}: The number of decisions that align with AI predictions, measured separately for the four cases before and after.
    \item \textbf{Decision Confidence}: Participants' confidence in their decision-making, rated on a 5-point Likert scale ranging from 1 (completely unsure) to 5 (extremely confident), also measured before and after across the four cases. 
  \end{enumerate}

After completing the decision-making task, participants filled in a post-experiment questionnaire to indicate their subjective opinions regarding the effectiveness of the explanations. This included satisfaction and perceived simplicity, as well as rankings of the explanations they relied on most. Participants’ educational background and experience with process mining or XAI were also collected to explore their potential influence on explanation effectiveness.





\subsection{Training Black-Box Model and Generating Explanations}
We use a real-life log to generate realistic explanations of different styles, train an outcome prediction model, and generate explanations. 
We select the BPIC 2017 event log~\cite{bpic17}, which is a real-life event log that records the loan application process of a Dutch financial institution. This dataset was selected for two primary reasons. First, it is a widely used dataset in the PPM and explainable PPM domains (e.g., ~\cite{DBLP:journals/tkdd/TeinemaaDRM19,DBLP:conf/caise/HundoganLDR23,DBLP:conf/caise/BuligaFGM23}), showing its applicability for generating process predictions and explanations, ensuring the feasibility of this study. Second, the loan application process provides a familiar and relatable context, making it accessible for participants, even those without prior knowledge of PPM. 

\begin{figure}[tb]
    \centering
    \includegraphics[width=\textwidth]{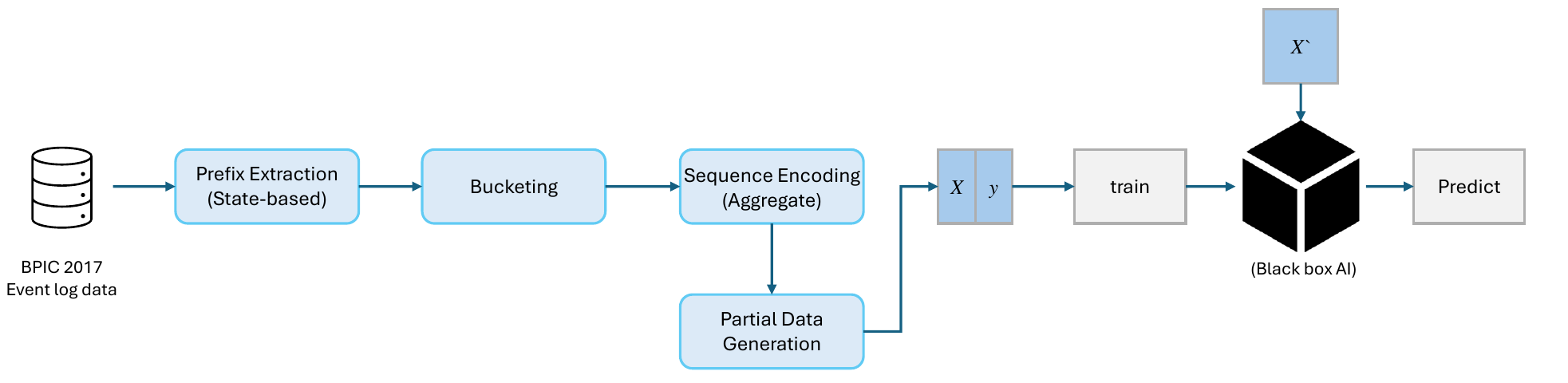}
    \caption{PPM workflow}
    \label{fig:xaiprocess}
\end{figure}

To prepare the data for model training, we followed established preprocessing steps~\cite{DBLP:journals/kbs/WickramanayakeH22,povalyaeva2017bpic}. Fig.~\ref{fig:xaiprocess} shows an overview of our PPM workflow. We used prefix extraction and state-based bucketing with the activity \textit{O\_Returned} as the mile-stone state, representing the decision point. We used aggregation encoding, counting all events up to the mile-stone state for each case, and extracting the relevant case attributes. 
Additionally, we generated three features (i.e., income, employment status, credit score, and age). Half of the feature values were randomly generated to introduce variability, independently of the outcome labels. The other half was generated based on the rules to ensure some useful features when generating explanations. 
We then trained a Random Forest model as the black-box model to predict the process outcome, determining whether a loan application would be accepted, canceled, or rejected. The model achieved an overall accuracy of 0.85. 

\begin{figure}[tb]
\begin{minipage}[b]{0.5\textwidth}
    \centering
    \includegraphics[width=\textwidth]{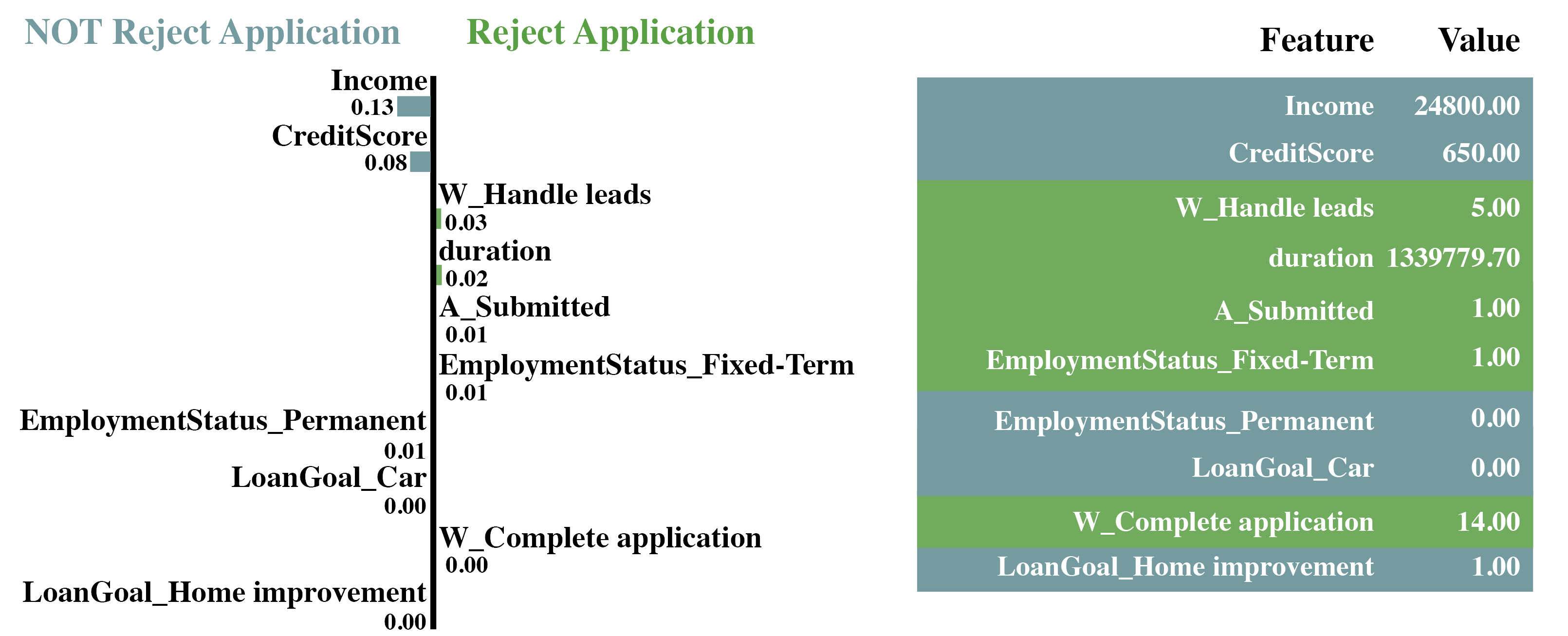}
    \caption{Feature example (LIME)}
    \label{fig:afteredit}
    \end{minipage}
\begin{minipage}[b]{0.5\textwidth}
    \centering
    \includegraphics[width=\textwidth]{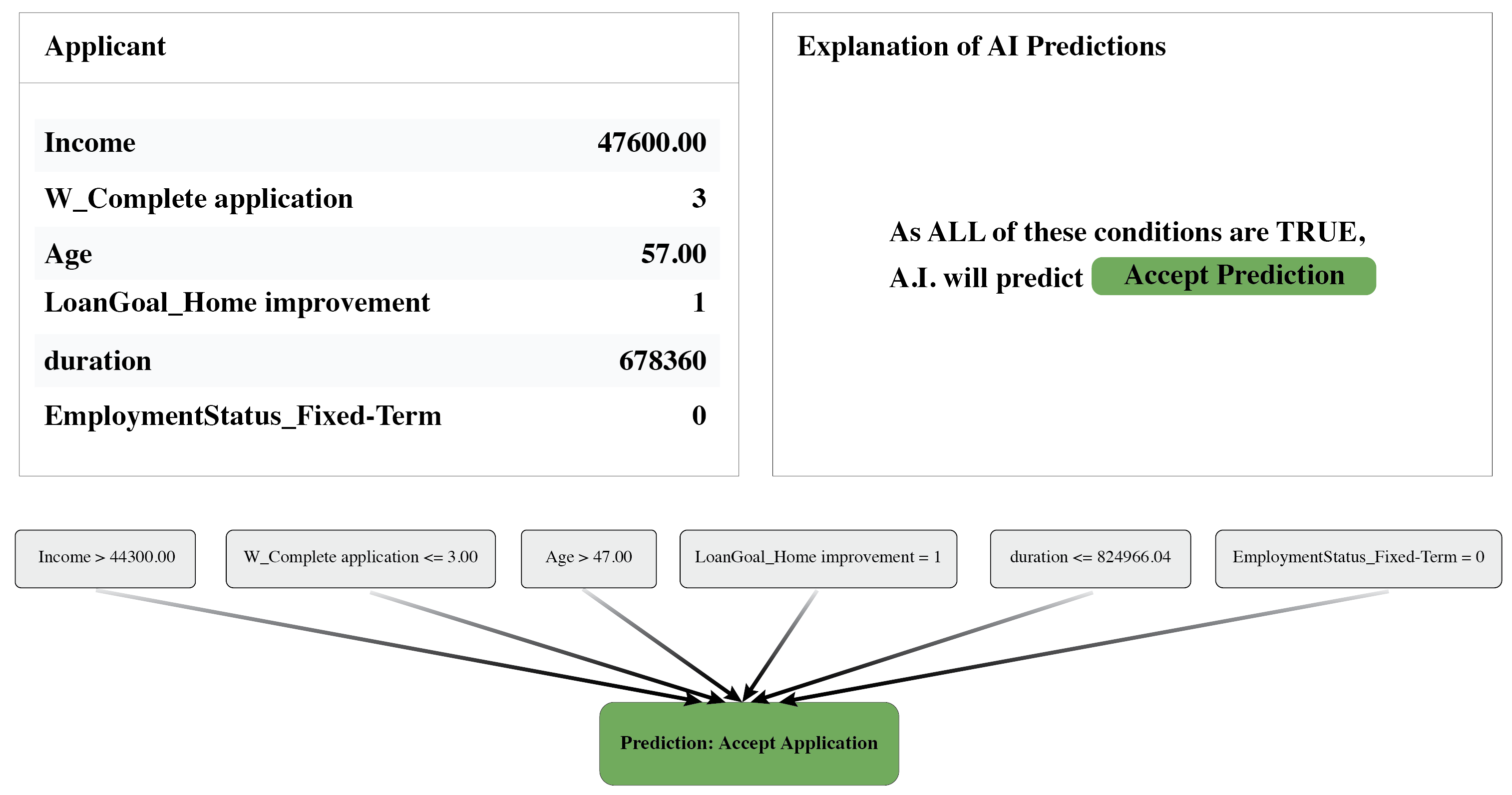}
    \caption{Rule example (Anchor)}
    \label{fig:afterule}
    \end{minipage}
    \end{figure}
\begin{figure}[tb]
    \centering
    \includegraphics[width=0.7\textwidth]{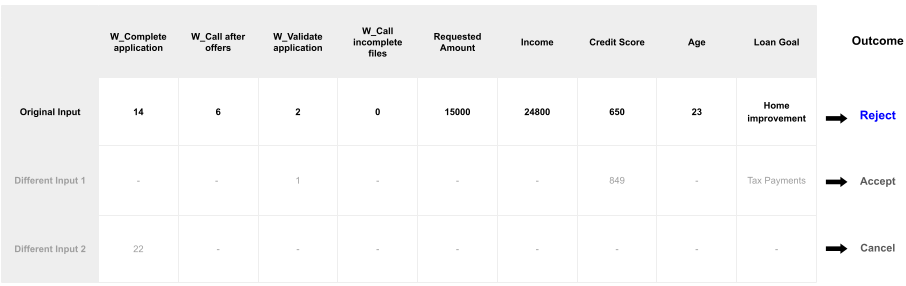}
    \caption{CF example (DiCE)}
    \label{fig:aftecf}
\end{figure}

To generate explanations for the black-box model, we employed three established XAI techniques, each of them representing a major explanation style used in PPM~\cite{DBLP:conf/icpm/HsiehMO21,DBLP:journals/corr/abs-2202-12018,DBLP:journals/algorithms/LeeCK22},:
\begin{itemize}
    \item Feature Importance (FI)-based explanations: LIME~\cite{DBLP:conf/kdd/Ribeiro0G16}, which identify variables influencing a prediction and highlight the most critical features.
    \item Rule-based explanations: Anchor~\cite{DBLP:conf/aaai/Ribeiro0G18}  which provide interpretable if-then rules that describe decision boundaries.
    \item Counterfactual-based explanations: DiCE~\cite{DBLP:conf/fat/MothilalST20}, which offer hypothetical scenarios to illustrate how a different outcome could occur under changed inputs.
\end{itemize}
Examples of each explanation style are are shown in Figures~\ref{fig:afteredit},~\ref{fig:afterule} and~\ref{fig:aftecf}. 
The full experimental setup, including detailed preprocessing steps and model configurations, is made available at~\url{https://github.com/ghksdl6025/evaluating_explanation_styles}.



\subsection{Survey Procedure} 

Before distributing the survey, we conducted pilot testing with five individuals, representing a mix of those with and without process mining knowledge and those with and without a Business IT background. This pilot test aimed to estimate the average time required to complete the task and assess the task difficulty. Participants without these backgrounds took approximately 20 minutes to complete the survey. Recognizing the varying levels of knowledge and anticipating that a long survey duration might reduce the quality of responses, we decided to reduce the number of tasks. Initially, the survey included five scenarios. 
Based on the pilot test feedback, we reduced this to four scenarios, focusing only on \emph{accept} and \emph{reject} predictions, each with one correct and one incorrect predictions. After these adjustments, the median completion time for the pilot survey was approximately 10 minutes.

The survey distribution was divided into multiple phases. The first batch consisted of approximately 20 Business Informatics master's students at Utrecht University attending a Process Mining lecture. The second batch included other master's students from backgrounds related to Business and IT, including Business Informatics. The third batch consisted of PhD researchers in the process mining field. The final batch was recruited from the Prolific platform\footnote{Copyright © [2024] Prolific. https://www.prolific.com} to fill in the required number of participants. In total, we gathered 222 participants. After excluding responses that are incomplete or the completion time under 5 minutes, we retained data from 179 participants. The median time to complete the survey for these qualified participants was 566.5 seconds, approximately 9-10 minutes. 

Participants were randomly assigned to one of six groups (approximately 30 participants per group) using the Qualtrics survey tool\footnote{Copyright © [2024] Qualtrics. https://www.qualtrics.com}. All participants followed the same procedure, which included the following stages:
\begin{enumerate}
    \item \textbf{Introduction}: Participants received an overview of the study objectives and the loan application process to provide context for the experiment.
    \item \textbf{Informed Consent}: A consent form was provided, detailing the study’s purpose, participants' rights, and data confidentiality measures.
    \item \textbf{Experiment}: Participants completed four decision-making tasks, making predictions both \emph{before} and \emph{after} receiving explanations.
    \item \textbf{Post Questionnaire}: Participants rated their satisfaction and their (perceived) simplicity with the explanations (1–5 scale). Open-ended responses captured additional feedback on challenges and suggestions for improvement.
    \item \textbf{Demographic Questionnaire}: Demographic data such as education level, STEM background, and experience with XAI and Process Mining were collected for analysis. Table~\ref{tab:demographic} summarizes the demographic distribution of the 179 participants.
\end{enumerate}

\begin{table}[!t]
\centering
\caption{Participant demographics (N=179)}
\label{tab:demographic}
\resizebox{0.6\textwidth}{!}{%
\begin{tabular}{llrr}
\hline
\multicolumn{2}{l}{\textbf{Characteristics}} & \textit{\textbf{N}} & \textbf{\%} \\ \hline
\multirow{4}{*}{Education (highest completed)} & High School & 3 & 1.7 \\
 & Bachelor or equivalent & 98 & 54.1 \\
 & Master or equivalent & 69 & 38.1 \\
 & Ph.D. or higher & 9 & 5 \\ \hline
\multirow{2}{*}{STEM Background} & Yes & 138 & 22.7 \\
 & No & 41 & 76.2 \\ \hline
\multirow{6}{*}{Process Mining Experience} & Never worked with it & 107 & 59.1 \\
 & Less than 1 year & 41 & 22.7 \\
 & 1-2 years & 17 & 9.4 \\
 & 2-3 years & 6 & 3.3 \\
 & 3-5 years & 4 & 2.2 \\
 & 5+ years & 4 & 2.2 \\ \hline
\multirow{6}{*}{XAI Experience} & Never worked with it & 107 & 59.1 \\
 & Less than 1 year & 32 & 17.7 \\
 & 1-2 years & 24 & 13.3 \\
 & 2-3 years & 6 & 3.3 \\
 & 3-5 years & 8 & 4.4 \\
 & 5+ years & 2 & 1.1 \\ \hline
\end{tabular}
}
\end{table}

An example of the full survey is publicly available\footnote{\label{code}see~\url{https://github.com/ghksdl6025/evaluating_explanation_styles}}. The survey is published at~\url{https://survey.uu.nl/jfe/form/SV_789q74Z184LzGJ0}. 
\section{Results}\label{sec:results}
This section presents the quantitative results obtained from the participants and details the analyses performed. The analyses were conducted using the Python Statsmodels package~\cite{seabold2010statsmodels}, which supports statistical tests. The code for the statistical tests is publicly available\footnotemark[\value{footnote}].

\subsection{Descriptive Statistics} 
Table~\ref{tab:batotal} provides a summary of the results for \emph{task performance}, \emph{agreement}, and \emph{decision confidence} across the three explanation styles and the two perceived accuracy levels.

\begin{table}[!b]
\centering
\caption{Before and After Explanation: Task Performance, Agreement, Decision Confidence among Accuracy and Exp. Styles}
\label{tab:batotal}
\resizebox{\textwidth}{!}{%
\begin{tabular}{cc|ccc|ccc|ccc}
\hline
\multirow{2}{*}{\textbf{Accuracy}} & \multirow{2}{*}{\textbf{\begin{tabular}[c]{@{}c@{}}Exp. \\ Styles\end{tabular}}} & \multicolumn{3}{c|}{\textbf{Task Performance}} & \multicolumn{3}{c|}{\textbf{Agreement}} & \multicolumn{3}{c}{\textbf{Decision Confidence}} \\ \cline{3-11} 
 &  & Before & After & Diff. & Before & After & Diff. & Before & After & Diff. \\ \hline
 & FI & 2.14 (0.68) & 2.21 (0.71) & 0.07 & 3.17 (1.02) & 2.90 (1.18) & -0.28 & 3.57 (0.88) & 3.61 (0.91) & 0.04 \\
High & Rule & 2.27 (0.89) & 2.23 (0.67) & -0.03 & 3.13 (0.96) & 3.30 (0.90) & 0.17 & 3.88 (0.63) & 4.07 (0.65) & 0.19 \\
 & CF & 1.70 (0.53) & 1.85 (0.59) & 0.15 & 3.59 (0.78) & 3.41 (0.78) & -0.19 & 4.02 (0.53) & 3.97 (0.54) & -0.05 \\ \hline
 & FI & 2.13 (0.66) & 2.45 (0.66) & 0.32 & 3.42 (0.79) & 3.16 (0.68) & -0.26 & 3.78 (0.93) & 3.87 (0.75) & 0.09 \\
Low & Rule & 2.39 (0.87) & 2.52 (0.88) & 0.13 & 3.29 (0.89) & 3.23 (0.97) & -0.06 & 3.73 (0.82) & 3.81 (0.84) & 0.08 \\
 & CF & 2.39 (0.66) & 2.84 (0.72) & 0.45 & 3.16 (0.88) & 3.48 (0.76) & 0.32 & 3.78 (0.54) & 3.66 (0.62) & -0.12 \\ \hline
\multicolumn{11}{l}{FI = Feature importance, CF = Counterfactual, Diff. = After - Before} \\ \hline
\end{tabular}%
}
\end{table}

As listed in Table~\ref{tab:batotal}, task performance varied within the high-accuracy group, depending on the type of explanation provided. The survey participants who received Feature importance (FI) explanations scored an average of 2.14 out of 4 before the explanation. This means that out of 4 decisions made before showing the explanation, 2.14 are made correctly on average across 29 participants (as the ground truth regardless of the prediction). This number slightly increased to 2.21 after showing the explanation. However, the standard deviations are rather high (0.68 before vs 0.71 after). 
Those who received Rule-based explanations scored 2.27 before and 2.23 after, slightly higher than FI. Those who received Counterfactual explanations had the lowest task performance in the group, with averages of 1.70 before and 1.85 after the explanation.

In the low-accuracy group, we observed the opposite for counterfactual, of which the task performance was the highest among the three groups, 2.39 before the explanation, and 2.84 after the explanation (with an increase of 0.45). 
Interestingly, participants in the low-accuracy group showed more noticeable improvements in their performance across all explanation styles after seeing the explanation. In the low-accuracy group, a mild increase of 0.32, 0.13, 0.45 respectively for FI, Rule, and CF, whereas only a slight increase of 0.07, -0.04, 0.15 in the high-accuracy group for the three styles. 

In the case of \emph{Agreement}, the results are more mixed, as listed in Column 5 and 6 in Table~\ref{tab:batotal}. 
The participants in the high-accuracy group who received Counterfactual explanations had the highest agreement with AI predictions, with 3.59 before and 3.41 after showing the explanation. 
In the low-accuracy group, participants with Counterfactual explanations showed an increase in agreement scores (from 3.16 to 3.48), reaching the highest agreement among the three styles. Another interesting observation is that after showing FI explanations,  the agreement score decreased in both high and low-accuracy groups, while for the rule-based explanation, the results are mixed. 

An interesting observation is that the average agreement score is consistently higher than the task performance, meaning out of the four cases, users agree more than three times with the prediction. Since two of these predictions are incorrect, these impact their task performance. This suggests that the users lost performance due to overreliance on the prediction.


Regarding \emph{Decision confidence} (see the last two columns in Table~\ref{tab:batotal}) scores varied across explanation styles and accuracy levels. In the high-accuracy group, Rule-based explanations led to the highest decision confidence after the explanation (4.07), followed by Counterfactual (3.97), and FI (3.61). In comparison, the low-accuracy group showed lower decision confidence scores, except for FI. 
When comparing the score before and after the explanation, the decision confidence increased for rule-based and FI, except for Counterfactual.


\subsection{Hypotheses Result} 


In this section, we present the results of our hypothesis testing. 
Table~\ref{tab:hypothesis_table} outlines the variables and statistical methods applied for each hypothesis. 

\begin{table}[b]
\centering
\caption{Statistical Analysis and Variables for Each Hypothesis}
\label{tab:hypothesis_table}
\resizebox{0.8\textwidth}{!}{%
\begin{tabular}{lll}
\hline
\multicolumn{1}{c}{\textbf{Independent Variable}} & \multicolumn{1}{c}{\textbf{Dependent Variable}} & \multicolumn{1}{c}{\textbf{Statistical test}} \\ \hline
\multicolumn{1}{l|}{\multirow{3}{*}{Perceived AI Acc. (\textbf{EXP1})}} & \multicolumn{1}{l|}{Before explanation TP (\textbf{H1})} & \multirow{3}{*}{Independent t-test} \\ \cline{2-2}
\multicolumn{1}{l|}{} & \multicolumn{1}{l|}{Before explanation Ag. (\textbf{H2})} &  \\ \cline{2-2}
\multicolumn{1}{l|}{} & \multicolumn{1}{l|}{Before explanation DC (\textbf{H3})} &  \\ \hline
\multicolumn{1}{l|}{\multirow{3}{*}{Exp. Styles (\textbf{EXP2})}} & \multicolumn{1}{l|}{Before vs. After explanation TP (\textbf{H4})} & \multirow{3}{*}{Paired t-test} \\ \cline{2-2}
\multicolumn{1}{l|}{} & \multicolumn{1}{l|}{Before vs. After explanation Ag. (\textbf{H5})} &  \\ \cline{2-2}
\multicolumn{1}{l|}{} & \multicolumn{1}{l|}{Before vs. After explanation DC (\textbf{H6})} &  \\ \hline
\multicolumn{3}{l}{TP = Task Performance, Ag. = Agreement, DC = Decision Confidence} \\ \hline
\end{tabular}
}
\end{table}

\subsubsection{EXP1: Effect of Perceived AI Accuracy on Initial Task Performance~(H1), Agreements~(H2), and Decision Confidence~(H3).}

We test the three hypotheses to measure the effect of perceived accuracy on decision-making in the before explanation setup. Independent sample t-tests were used for all three hypotheses. 

\myparagraph{\textbf{H1:} 
Effect of Perceived Accuracy on Task performance.}
H1 investigated whether perceived AI accuracy levels influence initial task performance. The independent variable is the accuracy group (high vs. low), and the dependent variable is the average task performance of the two groups prior to explanations.

The test yielded a p-value of 0.026 and an effect size of 0.336 (Cohen's d), indicating a small to medium effect. The results suggest a significant difference in task performance between the high-accuracy and low-accuracy groups. Interestingly, participants in the high-accuracy group performed \emph{worse} than those in the low-accuracy group, with a mean difference of -0.2558.  
This finding indicates that lower perceived AI accuracy is associated with \emph{higher initial task performance} before explanations are provided.


\myparagraph{\textbf{H2:} Effect of Perceived Accuracy on Agreement.} For H2, we investigated whether perceived AI accuracy influences participants’ agreement with AI predictions. The analysis showed no significant difference between the high-accuracy and low-accuracy groups, with a p-value of 0.998 and a negligible mean difference of 0.0349. These findings suggest that perceived AI accuracy does not significantly affect agreement with AI predictions in the pre-explanation setup.


\myparagraph{\textbf{H3:} Effect of Perceived Accuracy on Decision Confidence.} 
H3 evaluates the impact of perceived AI accuracy on participants' confidence in their decisions. The analysis yielded a p-value of 0.638, with a mean difference of 0.0698, indicating no significant difference in decision confidence between the two accuracy groups. These results suggest that perceived AI accuracy does not influence participants’ confidence in their decisions before explanations are provided.

\subsubsection{EXP2: Effect of Explanation Provision on Task Performance (H4), Agreements (H5), and Decision Confidence (H6)}


This section analyzes how the provision of explanations impacts decision-making, comparing pre- and post-explanation decisions. The hypotheses focus on how different explanation styles influence decision-making across high- and low-accuracy groups. Paired t-tests were conducted to assess these effects. The statistical results are summarized in Table~\ref{tab:effect_of_exp_stlyes}.


\begin{table}[b!]
\centering
\caption{P-Values (and Effect Sizes) for Paired t-Test Results: Changes in Task Performance, Agreement, and Decision Confidence}
\label{tab:effect_of_exp_stlyes}
\resizebox{0.8\textwidth}{!}{%
\begin{tabular}{lccc}
\hline
\multicolumn{1}{c}{\textbf{Target Groups}} & \textbf{\begin{tabular}[c]{@{}c@{}}Task performance\\ (H4)\end{tabular}} & \textbf{\begin{tabular}[c]{@{}c@{}}Agreement\\ (H5)\end{tabular}} & \textbf{\begin{tabular}[c]{@{}c@{}}Decision Confidence\\ (H6)\end{tabular}} \\ \hline
\textbf{Feature Importance (FI)} & 0.096 (0.218) & \cellcolor[HTML]{ffffbf}0.025 (0.296) & 0.396 (0.110) \\
\textbf{Rule-based (Rule)} & 0.689 (0.051) & 0.717 (0.047) & 0.135 (0.194) \\
\textbf{Counterfactual (CF)} & \cellcolor[HTML]{ffffbf}0.002 (0.425) & 0.497 (0.090) & 0.218 (0.164) \\
\textbf{High Acc. FI} & 0.663 (0.082) & 0.187 (0.251) & 0.655 (0.084) \\
\textbf{High Acc. Rule} & 0.839 (0.037) & 0.433 (0.145) & 0.076 (0.336) \\
\textbf{High Acc. CF} & 0.294 (0.206) & 0.259 (0.222) & 0.557 (0.114) \\
\textbf{Low Acc. FI} & 0.077 (0.329) & \cellcolor[HTML]{ffffbf}0.043 (0.379) & 0.476 (0.130) \\
\textbf{Low Acc. Rule} & 0.489 (0.126) & 0.712 (0.067) & 0.582 (0.100) \\
\textbf{Low Acc. CF} & \cellcolor[HTML]{ffffbf}0.002 (0.625) & 0.086 (0.318) & 0.285 (0.196) \\ \hline
\multicolumn{4}{l}{Note: \colorbox[HTML]{ffffbf}{Yellow} cells indicate significant results where p-values are below 0.05.} \\ \hline
\end{tabular}%
}
\end{table}

\myparagraph{\textbf{H4}: Effect of Explanation Provision on Task Performance. }
H4 examines whether providing explanations influences participants' task performance. The independent variable is the explanation provision, and the dependent variable is task performance.

Grouping the data by explanation styles (merging the low- and high-accuracy groups), the results of Counterfactual explanations show a significant increase in task performance. The paired t-test showed a t-value of -3.236 (df = 57), with p = 0.002 and a medium effect size of 0.425. The average increase in task performance is 0.310. For FI-based and Rule-based styles, no significant changes were observed. 

To confirm these results, we conducted further analysis, 
testing the high and low-accuracy groups separately (see the last six rows of Table~\ref{tab:effect_of_exp_stlyes}).
The test results show a significant improvement in task performance for the Counterfactual-based style, particularly in the low-accuracy group, with a p-value of 0.002 and a medium effect size of 0.625. This was also observed in the descriptive statistics in Table~\ref{tab:batotal}. 


These results allow us to reject the null hypothesis ($H4_0$), which posits that explanations do not influence task performance. The data support the alternative hypothesis ($H4_1$), indicating that \textbf{Counterfactual-based explanations significantly improve task performance}, particularly within the low-accuracy group, highlighted by a medium effect size.

\myparagraph{\textbf{H5}: Effect of Explanation Provision on Agreement.}
H5 investigates whether explanations influence participants' agreement with AI predictions. The independent variable is explanation provision, and the dependent variable is agreement before and after explanations.

Across all participants, the FI-based style showed a significant decrease in agreement, with a mean difference of 0.27 (p = 0.025) and a small to medium effect size of 0.296. No significant changes in agreement were observed for the Rule-based or Counterfactual-based styles, as their p-values exceeded 0.05.

When testing the high and low-accuracy groups separately, the FI-based style in the low-accuracy group remains significant, showing a decrease of -0.26 in agreement and a p-value of 0.043. The test results in the high-accuracy group became insignificant, with a p-value of 0.187. No significant changes in agreement were observed for the Rule-based or Counterfactual-based styles, as their p-values exceeded 0.05.
%
These findings suggest that \textbf{FI-based explanations may reduce agreement in low-accuracy scenarios}, but explanations generally have limited influence on agreement in high-accuracy scenarios.

\myparagraph{\textbf{H6}: Effect of Explanation Provision on Decision Confidence.}
H6 evaluates whether explanations influence participants' decision confidence. The independent variable is the explanation style, and the dependent variable is the change in confidence before and after explanations.

Across all participants, no significant changes in decision confidence were observed for any explanation style. The p-values were as follows: FI-based (p = 0.396), Rule-based (p = 0.135), Counterfactual-based (p = 0.218). Effect sizes were small, indicating negligible impact.

When analyzing the high and low accuracy groups separately, the results were non-significant for both groups: high accuracy (FI-based p=0.655, Rule-based p=0.076, Counterfactual-based p=0.557); low accuracy (FI-based p=0.476, Rule-based p=0.582, Counterfactual-based p= 0.285). Separate analysis for the high- and low-accuracy groups yielded non-significant results for all explanation styles. Based on these results, we fail to reject the null hypothesis ($H6_0$), which states that explanations do not influence participants' decision confidence.

\myparagraph{Satisfaction and Simplicity.}
Users' subjective evaluations of explanations regarding satisfaction and simplicity revealed that the Rule-based explanation style achieved the highest satisfaction scores and was rated the easiest to understand across both high- and low-accuracy groups (averages of 3.80 and 3.61, respectively). The Feature Importance (FI) style ranked second, with satisfaction scores of 3.66 and 3.39. In contrast, the Counterfactual explanation style received the lowest satisfaction ratings (3.41 and 3.19) and was perceived as slightly more difficult to understand. Interestingly, despite their effectiveness in enhancing task performance (as discussed previously), counterfactual explanations presented challenges in terms of user satisfaction and simplicity. 
These findings suggest that Rule-based explanations balance satisfaction and simplicity, while Counterfactual explanations may trade off user satisfaction and simplicity for performance gains.




\subsection{Discussion}

Interestingly, task performance was consistently better in the low-accuracy group compared to the high-accuracy group, even when the same explanation styles were applied. This counterintuitive result may indicate that users in the low-accuracy group engaged more critically with the provided explanations. 
With the high-accuracy label, users might have trusted the AI predictions too readily, relying less on the explanations to guide their decisions. This overreliance may have contributed to lower task performance in the high-accuracy group. This may suggest that it is important to foster appropriate skepticism among users when interacting with AI systems, especially in cases where predictions are uncertain or may be incorrect. 

Furthermore, the greater effect of explanations in the low-accuracy group (i.e., the improvement in task performance before and after the explanation) also aligns with the hypothesis that users are more likely to scrutinize explanations when they perceive the model as less reliable. In contrast, users in the high-accuracy group may view explanations as supplementary or only consult them to resolve doubts. This behavioral difference underscores the need to design explanations that can effectively support users in both scenarios, ensuring the explanations are used as intended.

Our findings also show that counterfactual explanations significantly enhance task performance, particularly in the low-accuracy group. This suggests that counterfactual-based explanations provide insights that enable users to more effectively identify incorrect decisions. 

It is important to note that the model’s accuracy was a label rather than an inherent characteristic of the underlying system, which remained the same across conditions. This design choice suggests that perceived accuracy significantly influences how users interact with explanations. The stronger impact of explanations in the low-accuracy group underscores the importance of managing user perceptions when presenting AI systems, particularly in contexts where interpretability and trust are critical. 

It is worth highlighting an important consideration regarding perceived AI accuracy. Initially, we explored the idea of using actual model performance by training two separate models—one with high accuracy and one with low accuracy. However, we found that the explanations generated by XAI techniques differed significantly between the two models. This shifted the focus of our study toward the content of the explanations (i.e., which features best explain the prediction) rather than the comparison of explanation styles. To ensure a fair comparison across groups, we chose to use a single model while manipulating only the perceived accuracy. 
Future work could investigate how true model accuracy affect user trust and decision-making.

\section{Conclusion}\label{sec:conclusion}
This study aimed to evaluate the \emph{effect} of explanation styles (i.e., Feature importance-based, Rule-based, and Counterfactual-based) and perceived AI accuracy on decision-making within the predictive process monitoring (PPM) domain. 
The evaluation was conducted in a decision-making scenario where participants assessed whether to accept or reject a loan application based on the AI's predictions. The effects were measured across three metrics: task performance, agreement, and decision confidence.

The findings revealed that perceived AI accuracy significantly influenced initial task performance, with lower perceived accuracy associated with higher initial task performance. Among the explanation styles, Counterfactual explanations proved to be particularly effective in enhancing task performance and agreement. In contrast, Feature importance explanations showed the lowest levels of agreement across all styles.

Rule-based explanations stood out in terms of user satisfaction and perceived simplicity, as they achieved the highest ratings. These characteristics also translated into the highest decision confidence. However, despite their positive reception, Rule-based explanations did not appear to significantly improve task performance. These results suggest there may be no correlation between the perceived satisfaction of an explanation style and its objective effectiveness in enhancing decision-making outcomes.

Building on these findings, future work could focus on conducting application-grounded evaluations involving domain experts to further validate the effectiveness of these explanation styles in real-world contexts. 
Additionally, future research could consider more advanced explainable predictive process monitoring techniques in the evaluation. 
Exploring novel explanation styles, such as human language-based explanations powered by Large Language Models (LLMs), represents another promising avenue for improving both interpretability and user engagement. 

\bibliographystyle{splncs04}
\bibliography{literature}

\end{document}